# Medical Image Segmentation and Localization using Deformable Templates

J.M.Spiller[1], T. Marwala[1]

[1] School of Electrical & Information Engineering, University of the Witwatersrand, Johannesburg, South Africa

*Abstract*— **This paper presents deformable templates as a tool for segmentation and localization of biological structures in medical images. Structures are represented by a prototype template, combined with a parametric warp mapping used to deform the original shape. The localization procedure is achieved using a multi-stage, multi-resolution algorithm designed to reduce computational complexity and time. The algorithm initially identifies regions in the image most likely to contain the desired objects and then examines these regions at progressively increasing resolutions. The final stage of the algorithm involves warping the prototype template to match the localized objects. The algorithm is presented along with the results of four example applications using MRI, x-ray and ultrasound images.**

## I. INTRODUCTION

Image segmentation and localization plays an important role in many medical imaging applications by automating or facilitating the delineation of anatomical structures. Since all subsequent interpretation tasks, such as feature extraction, object recognition, and classification depend largely on the quality of the segmentated output, effective segmentation has become a critical step for automated analysis in medical imaging. Segmentation and localization of anatomical structures is difficult in practice, especially when dealing with inherently noisy, low spatial resolution images such as those produced using functional imaging.

Deformable Templates provide a powerful tool for image segmentation, by exploiting constraints derived from the image data together with *a priori* knowledge about the location, size, and shape of the required structures [1].

The algorithm presented here is used to find a match between a deformed template and objects in the image, by minimizing a cost function between the template and object boundary. The algorithm achieves computational efficiency by searching the image in a number of stages and resolutions, refining the search at each stage. Object Localization requires that the template be matched regardless of the object's displacement, rotation, scale and deformation. Invariance to these characteristics is incorporated at each stage of the algorithm using discrete template orientations.

Localization of an object typically takes between 10s and 30s.

## II. A MODEL OF DEFORMATION

The model of deformation presented is based on the pattern theoretic model of Grenander [2]. It consists of two parts:
1. A prototype template image $T_0$ describing the overall architecture of the shape in terms of its representative contours and edges.
2. A set of control points (CPs) on the template, used to define a parametric mapping governing the deformation of the prototype.

The prototype template is based on average or expected evidence and is designed to capture expert, prior knowledge of the shape, size and orientation of the anatomical structure to be localized.

The CPs are defined on the image before the localization process commences and are used to facilitate warping of the prototype to create shape variation. These points can be placed to incorporate expert knowledge of where shape variation is more likely to occur.

This type of model is particularly appropriate in medical image segmentation, where inexact knowledge about the shape of an object is available and where it is necessary to accommodate the often significant variability of biological structures over time and across different individuals [3]. Figure 1 shows a typical template and base image.

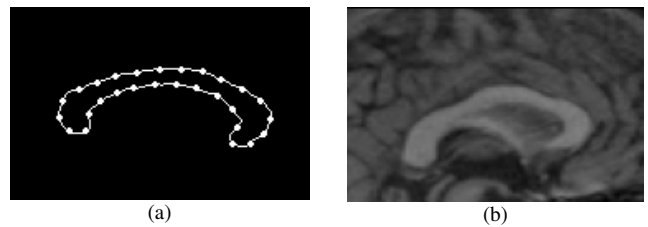

(a)          (b)

Fig. 1. Deformable Template Matching: (a) A prototype template of a typical Corpus Callosum shape with control points. (b) MRI base image where the Corpus Callosum must be localized and segmented.

## III. THE MULTI-STAGE ALGORITHM

In typical applications, an object has to be locatable regardless of translation, rotation and size. Given that the localization also needs to be invariant to partial shape changes, this introduces a large number of variables to be determined during optimization. Objective functions are typically non-convex, and so a multi-stage algorithm is employed to reduce the search space [1]. Using the algorithm described in this section, objects localization and segmentation is accomplished in 10s-30s.

*Stage 1 – Regions of Interest:* This stage is designed to reduce the search space by identifying the correct regions of the image to search for an object.

The template $T$ with dimensions $X$ and $Y$ is convolved with the edges of the base image $B$ using a 2D convolution filter in the spatial domain [4].

$$C(X,Y) = \sum_{x=1}^{size(X)} \sum_{y=1}^{size(Y)} T(x,y)B(X-x,Y-y) \quad (1)$$

The result of the convolution is given by Equation 1 and represents the image as an intensity map with high intensity where the convolution integral is large.

Template-sized regions of high intensity are searched for object matches in the next stage of the algorithm. Using only these areas, the visual search space can be reduced by up to 80%.

*Stage 2 – Multi-Resolution Approximate Matches:* During this stage, possible object matches are determined using approximate matching.

The template is windowed at discrete positions and orientations over the base image and a match between the two is evaluated. The discrete window locations do not cover the entire base image, but are obtained from Stage 1.

The template is attracted and aligned to the salient edges in the base image via directional Edge Potential Fields (EPFs), determined by the positions and directions of edges in the base image [1]. The EP for each pixel in the base image is defined by:

$$\Phi(x,y) = -\exp\left(-\sqrt{(\delta_x^2 + \delta_y^2)}\right) \quad (2)$$

Where $(\delta_x, \delta_y)$ is the displacement from the pixel to the nearest edge point in the image. A directional component is also included for each pixel $(x,y)$ in edge $I$, determined by:

$$\Theta(x,y) = \arctan\left(\frac{dI(x,y)}{dy} \bigg/ \frac{dI(x,y)}{dx}\right) \quad (3)$$

This modified EPF induces an energy function between the template image and the base image given by [1]:

$$E(T_{s,\Theta,\xi,d}, B) = \frac{1}{N_T} \sum_{i=1}^{N_T} \left(1 + \Phi_i(x,y)|\cos(\beta(x,y))|\right) \quad (4)$$

Where the summation is over the $N_T$ pixels on the template and $\beta(x, y)$ is the angle between the tangent of the edge pixel nearest $(x, y)$ and the tangent direction of the template at $(x, y)$. This energy measure, expressed as a correlation between the base image and the template deformed in terms of scale $s$, rotation $\Theta$, displacement $d$ and warp $\zeta$, is based on the Chamfer matching function but requires that the template agree with the image edges not only in position, but also in direction [1]. This requirement provides significantly improved robustness in noisy images. The locations of low energy matches (below an application-specific threshold) are taken as possible matches.

The resolution of the EPF image is controlled by varying the standard deviation σ, of the Gaussian filter used to find the image edges [5]. Once the set of possible matches is found at a coarse resolution Stage 2 is repeated at progressively finer resolutions to determine final matches. The locations of possible matches at previous resolutions are used as starting locations for finer resolution matching.

*Stage 3 – Template Deformation:* At this stage, the template is deformed to fit the image edges accurately. This deformation can be thought of as a registration between the template image and the base.

The approximate match locations from the finest resolution EPFs are used to initialize the template placement on the base edge map. Control points on the template are transferred to corresponding locations on the edge map and are then repositioned iteratively using Particle Swarm optimization (PSO) [6]. At each iteration of the PSO algorithm, a Local Weighted Mean (LWM) warp [7] is used to warp the template to fit these new CP positions. The PSO algorithm can be run for a specified number of iterations, or until a terminating criterion, such as a minimum energy, is met.

Given N corresponding CPs $(X_i, Y_i)$ on the template and $(x_i, y_i)$ on the base image, LWM warping requires two functions,

$$X_i \approx f(x_i, y_i)$$
$$Y_i \approx g(x_i, y_i)$$

that approximate a mapping between these points, as closely as possible. The transformation functions can be obtained directly from the given control points and do not require the solution of systems of equations. $f$ (and similarly $g$) are obtained from Equation 5 [7].

$$f(x,y) = \frac{\sum_{i=1}^{N} W\{[(x-x_i)^2 + (y-y_i)^2]^{\frac{1}{2}} / R_n\} P_i(x,y)}{\sum_{i=1}^{N} W\{[(x-x_i)^2 + (y-y_i)^2]^{\frac{1}{2}} / R_n\}} \quad (5)$$

Where the N polynomials $P_i$ are constructed using the Gram-Schmidt orthogonalization process, and a set of linearly independent functions $h_i$, to have the form [7]:

$$\begin{aligned}
P_0(x,y) &= a_{00} h_0(x,y) \\
P_1(x,y) &= a_{10} P_0(x,y) + a_{11} h_1(x,y) \\
P_2(x,y) &= a_{20} P_0(x,y) + a_{21} P_1(x,y) + a_{22} h_2(x,y) \\
&\vdots \\
P_T(x,y) &= a_{T0} P_0(x,y) + a_{T1} P_1(x,y) + \ldots + a_{TT} h_T(x,y)
\end{aligned} \quad (6)$$

Where *(x, y)* is any arbitrary point in the image. The N weight functions $W_i$ are given by:

$$\begin{aligned}
W_i(R) &= 1 - 3R^2 + 2R^3 & 0 \leq R \leq 1 \\
W_i(R) &= 0 & R > 1
\end{aligned} \quad (7)$$

Where $R = [(x-x_i)^2 + (y-y_i)^2]^{\frac{1}{2}} / R_n$ and $R_n$ is the distance of point *($x_i$, $y_i$)* from its (n-1)[th] nearest control point in the base image.

Apart from not requiring the solution of a system of equations at each iteration, this LWM warp implementation has a number of advantages over other warp methods[8]:

- Corresponding control points are not mapped exactly to each other so digital errors in the correspondences as well as small mismatch errors are smoothed.
- The rational weight functions adapt to the density and organization of the points and automatically extend to large gaps between control points.
- Because varied placement of control points is acceptable, expert knowledge of deformation can be incorporated.

The use of PSO to determine the optimal placement of CPs is effective because the regions of the global minima are known from Stage 2. To prevent the PSO from exploring outside of these regions, and to limit warp dimensionality, a penalty function is introduced that measures the sum of the squared differences between CPs on the original and deformed templates. The penalty function is added to the energy cost given by Equation 4 and penalizes extreme warps that would leave the region of global minimum. It is given by Equation 8, where $\alpha$ controls rigidity of the warp.

$$P(x,y) = \alpha \sum_{x=1}^{N} (X_i - x_i)^2 + (Y_i - y_i)^2 \quad (8)$$

IV. EXPERIMENTAL RESULTS

The deformable template model presented has been applied to different biological structures in a number of functional medical images.

The first test experiment presented involves the segmentation of the Corpus Callosum in four different MRI images. The prototype template used is the first Corpus Callosum shape. This experiment is designed to illustrate the warp capabilities of the algorithm, and the template image is initialized at the center of the base images. Figure 2 shows the initial and final base images. As can be seen, all four Corpus Callosums are localized and segmented, even though there is considerable shape variation between the images.

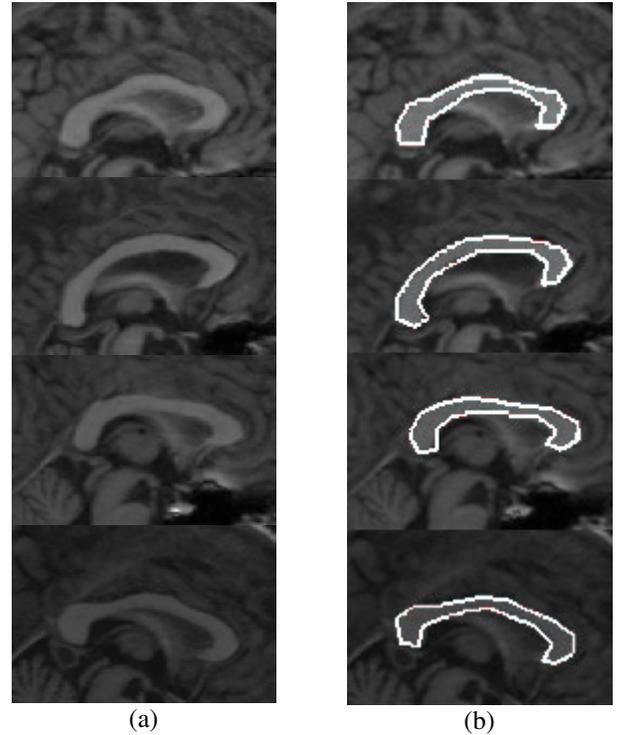

(a)        (b)

Fig. 2. Corpus Callosum MRI images. (a) Original image. (b) Segmented image.

The second experiment involves the detection of aneurysms in ultrasound images. The search algorithm is illustrated in this experiment, where an aneurysm is detected in two images regardless of template rotation and slight scale change. The two segmented ultrasound images are shown in Figure 3.

The third experiment involved the segmentation of cardiac MRI images. The two images of the heart were segmented using the same template, illustrating the warping capability as well as rotation invariance. This is shown in Figure 4.

achieve computation efficiency. The algorithm begins by identifying regions of interest in the image, and proceeds to search these regions at progressively finer resolutions. Once an object is located, the template is deformed to fit it using a Particle Swarm optimized, LWM warp routine. Experimental results have been presented showing invariant localization of objects in MRI, x-ray and ultrasound images.

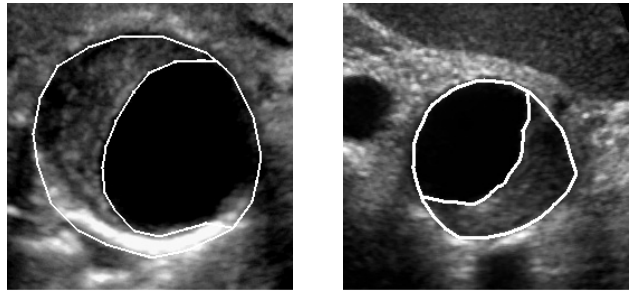

Fig. 3. Segmentation of Aneurysms in ultrasound images.

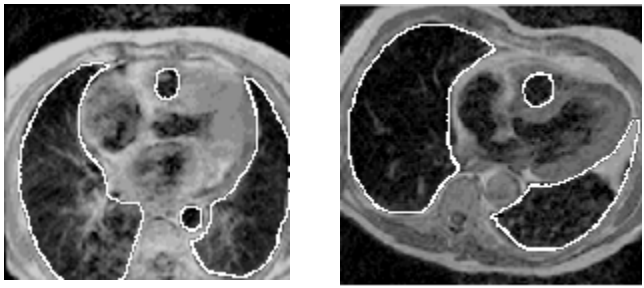

Fig. 4. Cardiac MRI segmentation.

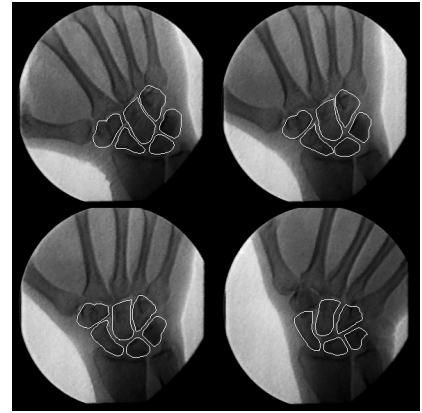

Fig. 5. Carpal Bone Segmentation from X-ray

The final experiment involves the detection of Carpal bones in x-ray images. This experiment shows how the algorithm can be adapted to object tracking tasks. X-ray images were taken of the hand and wrist moving in an ark. In each consecutive image, the final template from the previous image is used as the initial template for the current image. In this way, full localization is not required, resulting in speed and computational efficiency. Figure 5 shows the x-ray images, clockwise in consecutive order.

## V. CONLUSION

This paper presents a systematic approach to segmentation of medical images using deformable template. Prototype templates capture typical object structure, and are then used to localize similar structures within an image. The method utilizes a multi-stage, multi-resolution algorithm to

Address of the corresponding author:

J.M. Spiller
School of Electrical & Information Engineering
Private Bag 3, University of the Witwatersrand
Johannesburg, Wits 2050
South Africa
j.spiller@ee.wits.ac.za